\def\year{2017}\relax
\documentclass[letterpaper]{article}
\usepackage{aaai17}
\usepackage{times}
\usepackage{helvet}
\usepackage{courier}

\usepackage[pdftex]{graphicx}

\usepackage{amsmath}
\usepackage{amssymb}
\usepackage{color}
\usepackage{multirow}
\usepackage{url}            
\usepackage{booktabs}       

\allowdisplaybreaks

\begin{document}
%
\title{Growing Interpretable Part Graphs on ConvNets via Multi-Shot Learning}

\author{Quanshi Zhang, Ruiming Cao, Ying Nian Wu, and Song-Chun Zhu\\
University of California, Los Angeles}
\maketitle

\begin{abstract}
This paper proposes a learning strategy that extracts object-part concepts from a pre-trained convolutional neural network (CNN), in an attempt to 1) explore explicit semantics hidden in CNN units and 2) gradually grow a semantically interpretable graphical model on the pre-trained CNN for hierarchical object understanding. Given part annotations on very few (\emph{e.g.} 3--12) objects, our method mines certain latent patterns from the pre-trained CNN and associates them with different semantic parts. We use a four-layer And-Or graph to organize the mined latent patterns, so as to clarify their internal semantic hierarchy. Our method is guided by a small number of part annotations, and it achieves superior performance (about 13\%--107\% improvement) in part center prediction on the PASCAL VOC and ImageNet datasets\textcolor{red}{\footnote[1]{Codes here: \textit{https://sites.google.com/site/cnnsemantics/}}}.
\end{abstract}

\section{Introduction}

Convolutional neural networks~\cite{CNN,CNNImageNet,ResNet} (CNNs) have achieved near human-level performance in object classification on some datasets. However, in real-world applications, we are still facing the following two important issues.

Firstly, given a CNN that is pre-trained for object classification, it is desirable to derive an interpretable graphical model to explain explicit semantics hidden inside the CNN. Based on the interpretable model, we can go beyond the detection of object bounding boxes, and discover an object's latent structures with different part components from the pre-trained CNN representations.

Secondly, it is also desirable to learn from very few annotations. Unlike data-rich applications (\emph{e.g.} pedestrian and vehicle detection), many visual tasks demand for modeling certain objects or certain object parts \textit{on the fly}. For example, when people teach a robot to grasp the handle of a cup, they may not have enough time to annotate sufficient training samples of cup handles before the task. It is better to mine common knowledge of cup handles from a few examples on the fly.

Motivated by the above observations, in this paper, given a pre-trained CNN, we use very few (3--12) annotations to model a semantic part for the task of part localization. When a CNN is pre-trained using object-level annotations, we believe that its conv-layers have contained implicit representations of the objects. We call the implicit representations \textit{latent patterns}, each corresponding to a component of the semantic part (namely a sub-part) or a contextual region \emph{w.r.t.} the semantic part. \textit{For each semantic part, our goal is to mine latent patterns from the conv-layers related to this part. We use an And-Or graph (AOG) to organize the mined latent patterns to represent the semantic hierarchy of the part.}

\begin{figure}[t]
\centering
\includegraphics[width=0.99\linewidth]{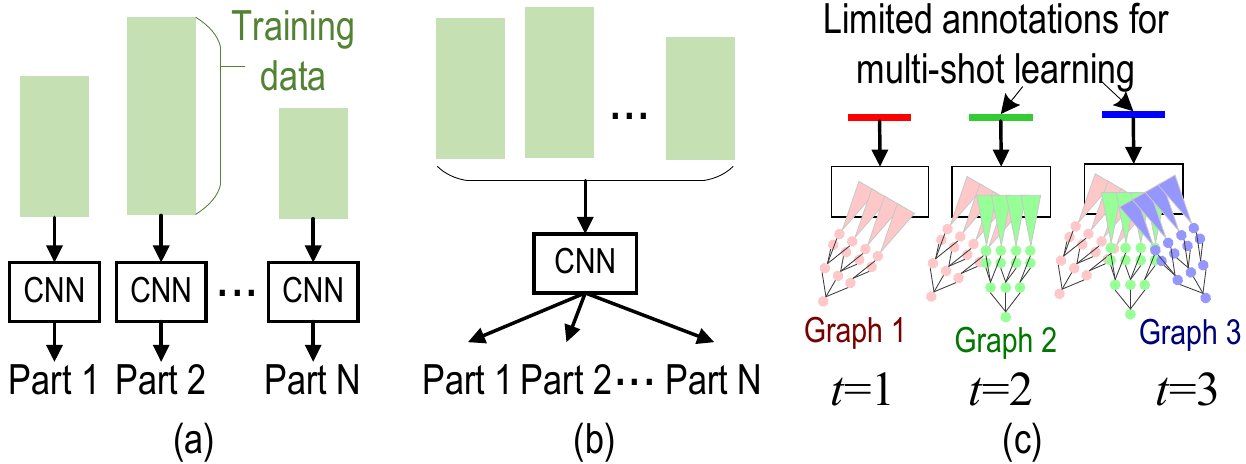}
\caption{Comparison of three learning strategies. (a) Individually learning/fine-tuning each part without sharing patterns between parts has large information redundancy in model representation. (b) Jointly learning/fine-tuning parts requires all parts to be simultaneously learned. (c) Given a small number (\emph{e.g.} 3--12) of part annotations based on demands on the fly, we incrementally grow new semantic graphs on a pre-trained CNN, which associate certain CNN units with new parts.}
\label{fig:top}
\end{figure}

\begin{figure*}[t]
\centering
\includegraphics[width=0.95\linewidth]{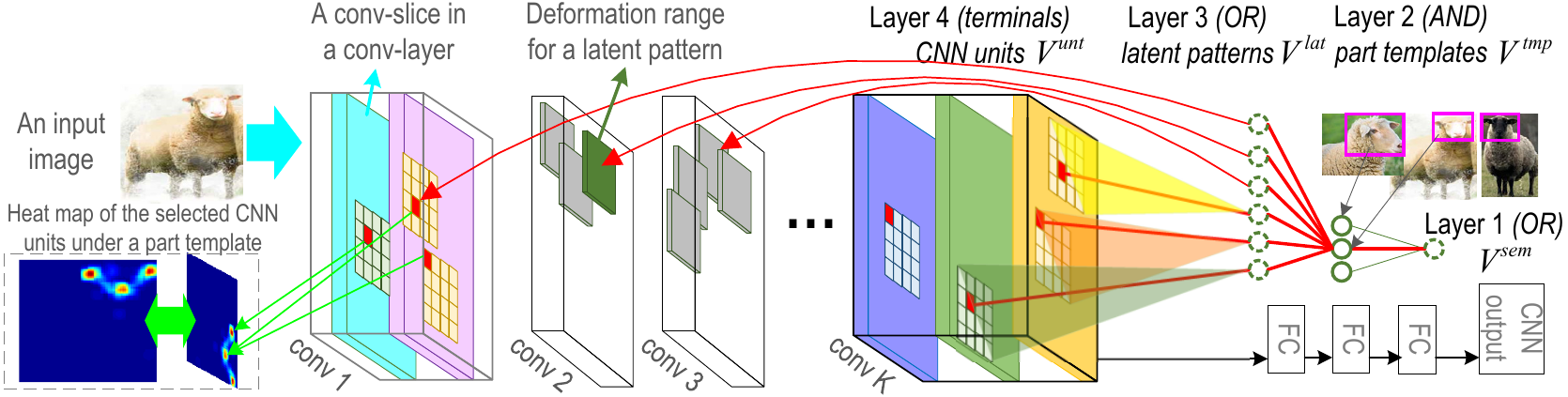}
\caption{Semantic And-Or graph grown on the pre-trained CNN. The AOG associates CNN units with certain semantic parts (head, here). Red lines in the AOG indicate a parse graph for concept association. To visualize these latent patterns, we show the heat map (left) at the 5-th conv-layers in the VGG-16 network, which sums up the associated units (red squares) throughout all conv-slices. In Fig.~\ref{fig:perf}, we reconstructed the dog head using the learned AOG to show its interpretability.}
\label{fig:model}
\end{figure*}

\textbf{Input and output:} Given a pre-trained CNN and a number of images for a certain category, we only annotate the semantic parts on a few images as input. We develop a method to grow a semantic And-or Graph (AOG) on the pre-trained CNN, which associates certain CNN units with the semantic part. Our method does not require massive annotations for learning, and can work with even a \textbf{single} part annotation. We can use the learned AOG to parse/localize object parts and their sub-parts for hierarchical object parsing.

Fig.~\ref{fig:model} shows that the AOG has four layers. In the AOG, each OR node encodes its alternative representations as children, and each AND node is decomposed into its constituents.
\begin{itemize}
\item Layer 1: the top OR node for \textit{semantic part} describes the head of a sheep in Fig.~\ref{fig:model}. It lists a number of part templates as children.
\item Layer 2: AND nodes for \textit{part templates} correspond to different poses or local appearances for the part, \emph{e.g.} a black sheep head from a front view and a white sheep head from side view.
\item Layer 3: OR nodes for \textit{latent patterns} describe sub-parts of the sheep head (\emph{e.g.} a corner of the nose) or a contextual region (\emph{e.g.} the neck region).
\item Layer 4: terminal nodes are \textit{CNN units}. A latent pattern naturally corresponds to a certain range of units within a conv-slice. It selects a CNN unit within this range to account for local shape deformation of this pattern.
\end{itemize}

\textbf{Learning method and key benefits:} The basic idea for growing AOG is to define a metric to distinguish reliable latent patterns from noisy neural activations in the conv-layers. We expect latent patterns with high reliability to 1) consistently represent certain sub-parts on the annotated object samples, 2) frequently appear in unannotated objects, and 3) keep stable spatial relationship with other latent patterns. We mine reliable latent patterns to construct the AOG. This learning method is related to previous studies of pursuing AOGs, which mined hierarchical object structures from Gabor wavelets on edges~\cite{MiningAoG} and HOG features~\cite{OurICCV15AoG}. We extend such ideas to feature maps of neural networks.

Our method has the following three key benefits:

\noindent$\bullet$ \textbf{CNN semanticization:} We semanticize the pre-trained CNN by connecting its units to an interpretable AOG. In recent years, people have shown a special interest in opening the black-box representation of the CNN. In this paper, we retrieve ``implicit'' patterns from the CNN, and use the AOG to associate each pattern with a certain ``explicit'' semantic part. We can regard the AOG as an interpretable representation of CNN patterns, which may contribute to the understanding of black-box knowledge organization in the CNN.

\noindent$\bullet$ \textbf{Multi-shot learning:} The idea of pattern mining also enables multi-shot learning from small data. Conventional end-to-end learning usually requires a large number of annotations to learn/finetune networks. In contrast, in our learning scenario, all patterns in the CNN have been well pre-trained using object-level annotations. We only use very few (3--12) part annotations to retrieve certain latent patterns, instead of finetuning CNN parameters. For example, we use the annotation of a specific tiger head to mine latent patterns. The mined patterns are not over-fitted to the head annotation, but represent common head appearance among different tigers. Therefore, we can greatly reduce the number of part annotations for training.

\noindent$\bullet$ \textbf{Incremental learning:} we can incrementally enrich the knowledge of semantic parts. Given a pre-trained CNN, we can incrementally grow new neural connections from CNN units to a new AOG, in order to represent a new semantic part. It is important to maintain the generality of the pre-trained CNN during the learning procedure. \emph{I.e.} we do not change/fine-tune the original convolutional weights within the CNN, when we grow new AOGs. This allows us to continuously add new semantic parts to the same CNN, without worrying about the model drift problem.

\textbf{Contributions} of this paper are summarized as follows.

\noindent
1) From the perspective of model learning, given a few part annotations, we propose a method to incrementally grow interpretable AOGs on a pre-trained CNN to gradually model semantic parts of the object.

\noindent
2) From the perspective of knowledge transferring, our method semanticizes a CNN by mining reliable latent patterns from noisy neural responses of the CNN and associating the implicit patterns with explicit semantic parts.

\noindent
3) To the best of our knowledge, we can regard our method as the first study to achieve weakly supervised (\emph{e.g.} 3--12 annotations) learning for part localization. Our method exhibits superior localization performance in experiments (about 13\%--107\% improvement in part center prediction).

\section{Related work}

\textbf{Long-term learning \& short-term learning:} As reported in \cite{ShortTermBrain}, there are ``two learning systems instantiated in mammalians:'' 1) the neocortex gradually acquires sophisticated knowledge representation, and 2) the hippocampus quickly learns specifics of individual experiences. CNNs are typically trained using big data, and contain rich appearance patterns of objects. If one compares CNNs to the neocortex, then the fast retrieval of latent patterns related to a semantic part can be compared to the short-term learning in hippocampus.

\textbf{Semantics in the CNN:} In order to explore the hidden semantics in the CNN, many studies have focused on the visualization of CNN units~\cite{CNNVisualization_1,CNNVisualization_2,CNNVisualization_3,CNNVisualization_4,CNNVisualization_5} and analyzed their statistical features~\cite{UnsuperCNN,CNNAnalysis_1,CNNAnalysis_2,CNNAnalysis_3}. Liu et al.~\cite{CNNFeatureMining} extracted and visualized a subspace of CNN features.


Going beyond ``passive'' visualization, some studies ``actively'' extracted CNN units with certain semantics for different applications. Zhou~\emph{et al.}~\cite{CNNSemanticDeep,CNNSemanticDeep2} discovered latent ``scene'' semantics from CNN feature maps. Simon~\emph{et al.} discovered objects~\cite{ObjectDiscoveryCNN_2} in an unsupervised manner from CNN feature maps, and learned semantic parts in a supervised fashion~\cite{CNNSemanticPart}. In our study, given very few part annotations, we mine CNN patterns that are related to the semantic part. Obtaining clear semantics makes it easier to transfer CNN patterns to other part-based tasks.

\textbf{AOG for knowledge transfer:} Transferring hidden patterns in the CNN to other tasks is important for neural networks. Typical research includes end-to-end fine-tuning and transferring CNN knowledge between different categories~\cite{CNNAnalysis_2,CNNSemantic} and/or datasets~\cite{UnsuperTransferCNN}. In contrast, we believe that a good explanation and transparent representation of part knowledge will creates a new possibility of transferring part knowledge. As in \cite{AllenAoG,MiningAoG}, the AOG is suitable to represent the semantic hierarchy, which enables semantic-level interactions between human and neural networks.

\textbf{Modeling ``objects'' vs. modeling ``\textbf{parts}'' in un-/weakly-supervised learning:} Generally speaking, in terms of un-/weakly-supervised learning, modeling parts is usually more challenging than modeling entire objects. Given image-level labels (without object bounding boxes), object discovery~\cite{ObjectDiscoveryCNN_1,ObjectDiscoveryCNN_2,ObjectDiscoveryCNN_3} can be achieved by identifying common foreground patterns from noisy background. Closed boundaries and common object structure are also strong prior knowledge for object discovery.

In contrast to objects, semantic parts are hardly distinguishable from other common foreground patterns in an unsupervised manner. Some parts (\emph{e.g.} the abdomen) do not have shape boundaries to determine their shape extent. Inspired by graph mining~\cite{OurICCV15AoG,OurSAPPAMI,OurCVPR14Graph}, we mine common patterns from CNN activation maps in conv-layers to explain the part.

\textbf{Part localization/detection vs. semanticizing CNN patterns:} Part localization/detection is an important task in computer vision~\cite{SSDPM,CNNSemanticPart,PLDPM,SemanticPart}. There are two key points to differentiate our study from conventional part-detection approaches. First, most methods for detection, such as the CNN and the DPM~\cite{CNN,SupervisedDPM_2,DeepPart}, limit their attention to the classification problem. In contrast, our effort is to clarify semantic meanings of implicit CNN patterns. Second, instead of summarizing knowledge from massive annotations, our method mines CNN semantics with very limited supervision.

\section{And-Or graph for part parsing}

In this section, we introduce the structure of the AOG and part parsing/localization based on the AOG. The AOG structure is suitable for clearly representing semantic hierarchy of a part. The method for mining latent patterns and building the AOG will be introduced in the next section. An AOG represents the semantic structure of a part at four layers.
\begin{small}
\begin{center}
\begin{tabular}{c|lcc}
\hline
Layer \!\!&\!\! Name \!\!&\!\! Node type \!\!&\!\! Notation\\
\hline
1 \!\!&\!\! semantic part \!\!&\!\! OR node \!\!&\!\! $V^{\textrm{sem}}$\\
2 \!\!&\!\! part template \!\!&\!\! AND node \!\!&\!\! $V^{\textrm{tmp}}\!\in\!\Omega^{\textrm{tmp}}$\\
3 \!\!&\!\! latent pattern \!\!&\!\! OR node \!\!&\!\! $V^{\textrm{lat}}\!\in\!\Omega^{\textrm{lat}}$\\
4 \!\!&\!\! CNN unit \!\!&\!\! Terminal node \!\!&\!\! $V^{\textrm{unt}}\!\in\!\Omega^{\textrm{unt}}$\\
\hline
\end{tabular}
\end{center}
\end{small}
Each OR node in the AOG represents a list of alternative appearance (or deformation) candidates. Each AND node is composed of a number of latent patterns to describe its sub-regions.

In Fig.~\ref{fig:model}, given CNN activation maps on an image $I$\textcolor{red}{\footnote[2]{Because the CNN has demonstrated its superior performance in object detection, we assume that the target object can be well detected by the pre-trained CNN. Thus, to simplify the learning scenario, we crop $I$ to only contain the object, resize it to the image size for CNN inputs, and only focus on the part localization task.}}, we can use the AOG for part parsing. From a top-down perspective, the parsing procedure 1) identifies a part template for the semantic part; 2) parses an image region for the selected part template; 3) for each latent pattern under the part template, it selects a CNN unit within a certain deformation range to represent this pattern.

In this way, we select certain AOG nodes in a \textit{parse graph} to explain sub-parts of the object (shown as red lines in Fig.~\ref{fig:model}). For each node $V$ in the parse graph, we parse an image region $\Lambda_{V}$ within image $I$\textcolor{red}{\footnote[3]{Image regions of OR nodes are propagated from their children. Each terminal node has a fixed image region, and each part template (AND node) has a fixed region scale (will be introduced later). Thus, we only need infer the center position of each part template in (\ref{eqn:AND}) during part parsing.}}. We use $S_{I}(V)$ to denote an inference/parsing score, which measures the fitness between the parsed region $\Lambda_{V}$ and $V$ (as well as the sub-AOG under $V$).

Given an image $I$\textcolor{red}{\footnotemark[2]} and an AOG, the actual parsing procedure is solved by dynamic programming in a bottom-up manner, as follows.

\textbf{Terminal nodes (CNN units):} We first focus on parsing configurations of terminal nodes. Terminal nodes under a latent pattern are displaced in location candidates of this latent pattern. Each terminal node $V^{\textrm{unt}}$ has a fixed image region {\small$\Lambda_{V^{\textrm{unt}}}$}: we propagate $V^{\textrm{unt}}$'s receptive field back to the image plane as {\small$\Lambda_{V^{\textrm{unt}}}$}. We compute $V^{\textrm{unt}}$'s inference score $S_{I}(V^{\textrm{unt}})$ based on both its neural response value and its displacement \emph{w.r.t.} its parent (see appendix for details\textcolor{red}{\footnote[4]{Please see the section of appendix for details.}}).

\textbf{Latent patterns:} Then, we propagate parsing configurations from terminal nodes to latent patterns. Each latent pattern $V^{\textrm{lat}}$ is an OR node. $V^{\textrm{lat}}$ naturally corresponds to a square within a certain conv-slice in the output of a certain CNN conv-layer as its deformation range\textcolor{red}{\footnote[5]{We set a constant deformation range for each latent pattern, which potentially covers {\small$75\!\times\!75$} pxls on the image. Deformation ranges of different patterns in the same conv-slice may overlap.}}. $V^{\textrm{lat}}$ connects all the CNN units within the deformation range as children, which represent different deformation candidates. Given parsing configurations of its children CNN units as input, $V^{\textrm{lat}}$ selects the child $\hat{V}^{\textrm{unt}}$ with the highest score as the true deformation configuration:
\begin{equation}
S_{I}(V^{\textrm{lat}})=\max_{V^{\textrm{unt}}\in Child(V^{\textrm{lat}})}S_{I}(V^{\textrm{unt}}), \quad\hat{\Lambda}_{V^{\textrm{lat}}}=\Lambda_{\hat{V}^{\textrm{unt}}}
\end{equation}

\textbf{Part templates:} Each part template $V^{\textrm{tmp}}$ is an AND node, which uses its children (latent patterns) to represent its sub-part/contextual regions. Based on the relationship between $V^{\textrm{tmp}}$ and its children, $V^{\textrm{tmp}}$ uses its children's parsing configurations to parse its own image region $\Lambda_{V^{\textrm{tmp}}}$. Given parsing scores of children, $V^{\textrm{tmp}}$ computes the image region $\hat{\Lambda}_{V^{\textrm{tmp}}}$ that maximizes its inference score.
\begin{equation}
S_{I}(V^{\textrm{tmp}})=\max_{\Lambda_{V^{\textrm{tmp}}}}\!\!\!\!\!\!\!\!\!\!\!\!\!\!\!\!\!\!\!\!\sum_{\quad\qquad V^{\textrm{lat}}\in\!Child(V^{\textrm{tmp}})}\!\!\!\!\!\!\!\!\!\!\!\!\!\!\!\!\!\!\!\!\!\big[S_{I}(V^{\textrm{lat}})+S^{\textrm{inf}}(\Lambda_{V^{\textrm{tmp}}}|\hat{\Lambda}_{V^{\textrm{lat}}})\big]\\
\label{eqn:AND}
\end{equation}
Just like typical part models (\emph{e.g.} DPMs), the AND node uses each child's region $V^{\textrm{lat}}$ to infer its own region. $S_{I}(V^{\textrm{lat}})$ measures the score of each child, and {\small$S^{\textrm{inf}}(\Lambda_{V^{\textrm{tmp}}}|\hat{\Lambda}_{V^{\textrm{lat}}})$} measures spatial compatibility between $V^{\textrm{tmp}}$ and each child $V^{\textrm{lat}}$ in region parsing (see the appendix for formulations).

\textbf{Semantic part:} Finally, we propagate parsing configurations to the top node $V^{\textrm{sem}}$. $V^{\textrm{sem}}$ is an OR node. It contains a list of alternative templates for the part. Just like OR nodes of latent patterns, $V^{\textrm{sem}}$ selects the child $\hat{V}^{\textrm{tmp}}$ with the highest score as the true parsing configuration:
\begin{equation}
S_{I}(V^{\textrm{sem}})\!=\!\max_{V^{\textrm{tmp}}\in Child(V^{\textrm{sem}})}S_{I}(V^{\textrm{tmp}}), \quad\hat{\Lambda}_{V^{\textrm{sem}}}\!=\!\hat{\Lambda}_{\hat{V}^{\textrm{tmp}}}\!\!
\end{equation}


\section{Learning: growing an And-Or graph}
\label{sec:AOGBuild}

The basic idea of AOG growing is to distinguish reliable latent patterns from noisy neural responses in conv-layers and use reliable latent patterns to construct the AOG.

\textbf{Training data:}{\verb| |} Let ${\bf I}$ denote an image set for a target category. Among all objects in ${\bf I}$, we label bounding boxes of the semantic part in a small number of images, ${\bf I}^{\textrm{ant}}\!=\!\{I_1,I_2,\ldots,I_{M}\}\subset{\bf I}$. In addition, we manually define a number of templates for the part. Thus, for each {\small$I\in{\bf I}^{\textrm{ant}}$}, we annotate {\small$(\Lambda_{V^{\textrm{sem}}}^{*},V^{\textrm{tmp}*})$}, where {\small$\Lambda_{V^{\textrm{sem}}}^{*}$} denotes the ground-truth bounding box of the part in $I$, and $V^{\textrm{tmp}*}$ specifies the ground-truth template ID for the part.

\textbf{Which AOG parameters to learn:}{\verb| |} We can use human annotations to define the first two layers of the AOG. If people specify a total of $m$ different part templates during the annotation process, correspondingly, we can directly connect the top node with $m$ part templates $\{V^{\textrm{tmp}*}\}$ as children. For each part template {\small$V^{\textrm{tmp}}$}, we fix a constant scale for its region $\Lambda_{V^{\textrm{tmp}}}$. \emph{I.e.} if there are $n$ ground-truth part boxes that are labeled for {\small$V^{\textrm{tmp}}$}, we compute the average scale among the $n$ part boxes as the constant scale for $\Lambda_{V^{\textrm{tmp}}}$.

Thus, the key to AOG construction is to mine children latent patterns for each part template. We need to mine latent patterns from a total of $K$ conv-layers. We select $n_{k}$ latent patterns from the $k$-th ($k=1,2,\ldots,K$) conv-layer, where $K$ and $\{n_{k}\}$ are hyper-parameters. Let each latent pattern {\small$V^{\textrm{lat}}$} in the $k$-th conv-layer correspond to a square deformation range\textcolor{red}{\footnotemark[5]}, which is located in the {\small$D_{V^{\textrm{lat}}}$}-th conv-slice of the conv-layer. {\small$\overline{\bf P}_{V^{\textrm{lat}}}$} denotes the center of the range. As analyzed in the appendix, we only need to estimate the parameters of {\small$D_{V^{\textrm{lat}}},\overline{\bf P}_{V^{\textrm{lat}}}$} for {\small$V^{\textrm{lat}}$}.

\textbf{How to learn:}{\verb| |} We mine the latent patterns by estimating their best locations {\small$D_{V^{\textrm{lat}}},\overline{\bf P}_{V^{\textrm{lat}}}\in{\boldsymbol\theta}$} that maximize the following objective function.
\begin{small}
\begin{equation}
\begin{split}
&\underset{\boldsymbol\theta}{\max}\!\Big\{\!\underbrace{\!\underset{I\in{\bf I}^{\textrm{ant}}}{\textrm{mean}}\!\Big[S_{I}(V^{\textrm{sem}})-\lambda_{V^{\textrm{tmp}*}}\Vert\hat{\bf P}_{V^{\textrm{sem}}}-{\bf P}^{*}_{V^{\textrm{sem}}}\Vert\Big]}_{\textrm{annotated images}}\\
&\qquad+\!\underbrace{\underset{I'\in{\bf I}}{\textrm{mean}}\sum_{V^{\textrm{lat}}}\!S^{\textrm{unsup}}_{I'}(V^{\textrm{lat}})}_{\textrm{unannotated images}}\!\Big\}\!\!\!
\end{split}
\label{eqn:build}
\end{equation}
\end{small}
where {\small${\boldsymbol\theta}$} is the set of AOG parameters. First, let us focus on the first half of the equation, which learns from part annotations. Given annotations $(\Lambda_{V^{\textrm{sem}}}^{*},V^{\textrm{tmp}*})$ on $I$, {\small$S_{I}(V^{\textrm{sem}})$} denotes the parsing score of the part. {\small$\Vert\hat{\bf P}_{V^{\textrm{sem}}}-{\bf P}^{*}_{V^{\textrm{sem}}}\Vert$} measures localization error between the parsed part region $\hat{{\bf P}}_{V^{\textrm{sem}}}$ and the ground truth ${\bf P}^{*}_{V^{\textrm{sem}}}$. We ignore the small probability of the AOG assigning an annotated image with an incorrect part template to simplify the computation of parsing scores, \emph{i.e.} {\small$S_{I}(V^{\textrm{sem}})\approx S_{I}(V^{\textrm{tmp}*})$}.

The second half of (\ref{eqn:build}) learns from objects without part annotations. We formulate {\small$S^{\textrm{unsup}}_{I'}\!(V^{\textrm{lat}})\!=\!\lambda^{\textrm{unsup}}\!\big[\!S_{I'}^{\textrm{rsp}}(\hat{V}^{\textrm{unt}})\!+\!S_{I'}^{\textrm{loc}}\!(V^{\textrm{unt}})\!-\!\lambda^{\textrm{close}}\!\Vert\Delta{\bf P}_{V^{\textrm{lat}}}\Vert^2\!\big]$}, where latent pattern {\small$V^{\textrm{lat}}$} selects CNN unit {\small$\hat{V}^{\textrm{unt}}$} as its deformation configuration on $I'$. The first term {\small$S_{I'}^{\textrm{rsp}}(\hat{V}^{\textrm{unt}})$} denotes the neural response of the CNN unit {\small$\hat{V}^{\textrm{unt}}$}. The second term {\small$S_{I'}^{\textrm{loc}}(V^{\textrm{unt}})=-\lambda^{\textrm{loc}}\Vert\hat{\bf P}_{V^{\textrm{unt}}}-\overline{\bf P}_{V^{\textrm{lat}}}\Vert^2$} measures the deformation level of the latent pattern. The third term measures the spatial closeness between the latent pattern and its parent $V^{\textrm{tmp}}$. We assume that 1) latent patterns that frequently appear among unannotated objects may potentially represent stable sub-parts and should have higher priorities; and that 2) latent patterns spatially closer to $V^{\textrm{tmp}}$ are usually more reliable. Please see the appendix for details of {\small$S_{I'}^{\textrm{rsp}}(\hat{V}^{\textrm{unt}})$} and scalar weights of $\lambda^{\textrm{unsup}}$, $\lambda^{\textrm{close}}$, and $\lambda^{\textrm{loc}}$.

When we set $\lambda_{V^{\textrm{tmp}*}}$ to a constant $\lambda^{\textrm{inf}}\sum_{k=1}^{K}n_{k}$, we can transform the learning objective in (\ref{eqn:build}) as follows.
\begin{small}
\begin{equation}
\forall V^{\textrm{tmp}}\in\Omega^{\textrm{tmp}},\quad\max_{{\boldsymbol\theta}_{V^{\textrm{tmp}}}}{\bf L},\quad {\bf L}\!=\!\!\!\!\!\!\!\!\sum_{V^{\textrm{lat}}\in Child(V^{\textrm{tmp}})}\!\!\!\!\!\!\!Score(V^{\textrm{lat}})
\label{eqn:subAOG}
\end{equation}
\end{small}
where {\small$Score(V^{\textrm{lat}})\!=\!{\textrm{mean}}_{I\in{\bf I}_{V^{\textrm{tmp}}}}[S_{I}(V^{\textrm{lat}})+S^{\textrm{inf}}(\Lambda^{*}_{V^{\textrm{sem}}}|\hat{\Lambda}_{V^{\textrm{lat}}})]$
$+{\textrm{mean}}_{I'\in{\bf I}}S^{\textrm{unsup}}_{I'}(V^{\textrm{lat}})$}. ${\boldsymbol\theta}_{V^{\textrm{tmp}}}\subset{\boldsymbol\theta}$ denotes the parameters for the sub-AOG of $V^{\textrm{tmp}}$. We use {\small${\bf I}_{V^{\textrm{tmp}}}\subset{\bf I}^{\textrm{ant}}$} to denote the subset of images that are annotated with $V^{\textrm{tmp}}$
as the ground-truth part template.

\textbf{Learning the sub-AOG for each part template:}{\verb| |} Based on (\ref{eqn:subAOG}), we can mine the sub-AOG for each part template $V^{\textrm{tmp}}$, which uses this template's own annotations on images $I\in{\bf I}_{V^{\textrm{tmp}}}\subset{\bf I}^{\textrm{ant}}$, as follows.

\noindent
1) We first enumerate all possible latent patterns corresponding to the $k$-th CNN conv-layer ($k=1,\ldots,K$), by sampling all pattern locations \emph{w.r.t.} $D_{V^{\textrm{lat}}}$ and $\overline{\bf P}_{V^{\textrm{lat}}}$.

\noindent
2) Then, we sequentially compute {\small$\hat{\Lambda}_{V^{\textrm{lat}}}$} and {\small$Score(V^{\textrm{lat}})$} for each latent pattern.

\noindent
3) Finally, we sequentially select a total of $n_{k}$ latent patterns. In each step, we select {\small$\hat{V}^{\textrm{lat}}\!=\!{\arg\!\max}_{V^{\textrm{lat}}}\Delta{\bf L}$}. \emph{I.e.} we select latent patterns with top-ranked values of {\small$Score(V^{\textrm{lat}})$} as {\small$V^{\textrm{tmp}}$}'s children.

\section{Experiments}

\subsection{Implementation details}

We chose the 16-layer VGG network (VGG-16)~\cite{VGG} that was pre-trained using the 1.3M images in the ImageNet ILSVRC 2012 dataset~\cite{ImageNet} for object classification. Then, given a target category, we used images in this category to fine-tune the original VGG-16 (based on the loss for classifying target objects and background). VGG-16 has 13 conv-layers and 3 fully connected layers. We chose the last 9 (from the 5-th to the 13-th) conv-layers as valid conv-layers, from which we selected units to build the AOG.

Note that during the learning process, we applied the following two techniques to further refine the AOG model. First, multiple latent patterns in the same conv-slice may have similar positions {\small$\overline{\bf P}_{V^{\textrm{lat}}}$}, and their deformation ranges may highly overlap with each other. Thus, we selected the latent pattern with the highest $Score(V^{\textrm{lat}})$ within each small range of $\epsilon\times\epsilon$ in this conv-slice, and removed other nearby patterns to obtain a spare AOG structure. Second, for each {\small$V^{\textrm{tmp}}$}, we estimated $n_{k}$, \emph{i.e.} the best number of latent patterns in conv-layer $k$. We assumed that scores of all the latent patterns in the $k$-th conv-layer follow the distribution of {\small$Score(V^{\textrm{lat}})\sim\alpha\exp[-(\beta{rank})^{0.5}]+\gamma$}, where $rank$ denotes the score rank of {\small$V^{\textrm{lat}}$}. We found that when we set {\small$n_{k}=\lceil0.5/\beta\rceil$}, the AOG usually had reliable performance.

\subsection{Datasets}

\begin{table}[t]
\centering\begin{small}
\caption{Average number of children}
\resizebox{1.0\linewidth}{!}{\begin{tabular}{p{2cm}|ccc}
\hline
\!\!\!{AOG Layer}\!\!\!\!\!\!&\!\!\!\!\!\! \;\;\;\#1 semantic \!\!\!\!\!\!&\!\!\!\!\!\! \;\;\;\#2 part \!\!\!\!\!\!&\!\!\!\!\!\! \;\;\;\#3 latent \!\!\!\\
&\!\!\! part\!\!\!\!\!\!&\!\!\!\!\!\! template \!\!\!\!\!\!&\!\!\!\!\!\! pattern\!\!\!\\
\hline
\!\!\!{Children number}\!\!\!&\!\!\! {3} \!\!\!&\!\!\! {4575.8} \!\!\!&\!\!\! {136.4} \!\!\!\\
\hline
\end{tabular}}
\label{tab:stat}
\end{small}
\end{table}

We tested our method on three benchmark datasets: the PASCAL VOC Part Dataset~\cite{SemanticPart}, the CUB200-2011 dataset~\cite{CUB200}, and the ILSVRC 2013 DET dataset~\cite{ImageNet}. Just like in most part-localization studies~\cite{SemanticPart}, we also selected six animal categories---\textit{bird, cat, cow, dog, horse}, and \textit{sheep}---from the PASCAL Part Dataset for evaluation, which prevalently contain non-rigid shape deformation. The CUB200-2011 dataset contains 11.8K images of 200 bird species. As in \cite{ActivePart,CNNSemanticPart}, we regarded these images as a single bird category by ignoring the species labels. All the above seven categories have ground-truth annotations of the \textit{head} (it is the \textit{forehead} part in the CUB200-2011 dataset) and \textit{torso/back}. Thus, for each category, we learned two AOGs to model its \textit{head} and \textit{torso/back}, respectively.

In order to provide a more comprehensive evaluation of part localization, we built a larger object-part dataset based on the off-the-shelf ILSVRC 2013 DET dataset. We used 30 \textbf{animal} categories among all the 200 categories in the ILSVRC 2013 DET dataset. We annotated bounding boxes for the \textit{heads} and \textit{front legs/feet} in these animals as two common semantic parts for evaluation. In Experiments, we annotated 3--12 boxes for each part to build the AOG, and we used the rest images in the dataset as testing images.

\subsection{Two experiments on multi-shot learning}

We applied our method to all animal categories in the above three benchmark datasets. We designed two experiments to test our method in the scenarios of {\small$(1\!\times\!3)$}-shot learning and {\small$(4\!\times\!3)$}-shot learning, respectively. We applied the learned AOGs to part localization for evaluation.

\textbf{Exp.~1, three-shot AOG construction:}{\verb| |} For each semantic part of an object category, we learn three different part templates. We annotated a single bounding box for each part template. Thus, we used a total of three annotations to build the AOG for this part.

\textbf{Exp.~2, AOG construction with more annotations:}{\verb| |} We continuously added more part annotations to check the performance changes. Just as in Experiment 1, each part contains the same three part templates. For each part template, we annotated four parts in four different object images to build the corresponding AOG.

\subsection{Baselines}

We compared our method with the following nine baselines. The first baseline was the fast-RCNN~\cite{FastRCNN}. We directly used the fast-RCNN to detect the target parts on objects. To enable a fair comparison, we learned the fast-RCNN by first fine-tuning the VGG-16 network of the fast-RCNN using all object images in the target category and then training the fast-RCNN using the part annotations. The second baseline was the strongly supervised DPM (SS-DPM)~\cite{SSDPM}, which was trained with part annotations for part localization. The third baseline was proposed in \cite{PLDPM}, which trained a DPM component for each object pose to localize object parts (namely, PL-DPM). We used the graphical model proposed in \cite{SemanticPart} as the fourth baseline for part localization (PL-Graph). The fifth baseline, namely CNN-PDD, was proposed by \cite{CNNSemanticPart}, which selected certain conv-slices (channels) of the CNN to represent the target object part. The sixth baseline (VGG-PDD-finetuned) was an extension of CNN-PDD, which was conducted based the VGG-16 network that was pre-fine-tuned using object images in the target category. Because in the scope of weakly supervised learning, ``simple'' methods are usually insensitive to the over-fitting problem, we designed the last three baselines as follows. Given the pre-trained VGG-16 network that was used in our method, we directly used this network to extract \textit{fc7} features from image patches of the annotated parts, and learned a linear SVM and a RBF SVM to classify target parts and background. Then, given a testing image, the three baselines brutely searched part candidates from the image, and used the linear SVM, the RBF SVM, and the nearest-neighbor strategy, respectively, to detect the best part. All the baselines were conducted using the same set of annotations for a fair comparison.

\begin{table}[t]
\begin{center}
\caption{Part localization performance}\label{tab:IoU}
\resizebox{1.0\linewidth}{!}{\begin{tabular}{p{2cm}|cccc|ccc}
\hline
& \multicolumn{4}{|c|}{{\bf Exp.~1}: {\bf 3-shot} learning} & \multicolumn{3}{|c}{{\bf Exp.~2}: {\bf 12-shot} learning}\\
\hline
Dataset & \multicolumn{2}{|c|}{Pascal} & \multicolumn{2}{|c|}{ImageNet} & \multicolumn{2}{|c|}{Pascal} & \multicolumn{1}{|c}{ImageNet}\\
\hline
Semantic part \!\!\!&\!\!\! Head \!\!\!&\!\!\! Torso \!\!\!& \multicolumn{1}{|c}{Head} &\!\!\! {\footnotesize F-legs} \!\!\!&\!\!\! Head \!\!\!&\!\!\! Torso \!\!\!& \multicolumn{1}{|c}{Head}\\
\hline
SS-DPM & 1.5 & 7.8 & 7.0 & 5.1 & 3.4 & 10.8 & 18.8\\
PL-DPM & 1.1 & 2.1 & 3.9 & 1.2 & 3.1 & 4.2 & 6.5\\
PL-Graph & 8.6 & 19.8 & 15.9 & 6.3 & 7.0 & 23.5 & 25.1\\
Fast-RCNN & 15.5 & 40.0 & 31.9 & 10.9 & 31.3 & 52.9 & 53.6\\
Ours & {\bf 29.2} & {\bf 50.7} & {\bf 40.1} & {\bf 23.2} & {\bf 34.6} & {\bf 59.5} & {\bf 55.0}\\
\hline
\end{tabular}}
\end{center}
Because object parts may not have clear boundaries, many studies~\cite{ObjectDiscoveryCNN_1,CNNSemanticPart} did not consider part scales in evaluation. Similarly, our method mainly localizes part center, and does not discriminatively learn a model for the regression of part bounding boxes, which is different from fast-RCNN methods. Instead, we simply fix a constant bounding-box scale for each part template, \emph{i.e.} the average scale of part annotations for this part template. Nevertheless, our method still exhibits superior performance.
\end{table}

\begin{table}[t]
\caption{Normalized distance of part localization. The performance was evaluated using the CUB200-2011 dataset.}
\label{tab:dist}
\centering
\resizebox{1.0\linewidth}{!}{\begin{tabular}{p{2.7cm}|cc|cc}
\hline
& \multicolumn{2}{|c|}{\footnotesize{\bf Exp.~1}: {\bf 3-shot} learning} & \multicolumn{2}{|c}{\footnotesize{\bf Exp.~2}: {\bf 12-shot} learning}\\
\hline
Semantic part & Head & Torso & Head & Torso\\
\hline
SS-DPM & 0.3469 & 0.2604 & 0.2925 & 0.2427\\
PL-DPM & 0.3412 & 0.2329 & 0.3056 & 0.1998\\
PL-Graph & 0.4889 & 0.4015 & 0.6093 & 0.3961\\
\!\!\!{\footnotesize fc7+linearSVM}\!\!\! & 0.3120 & 0.2721 & 0.2906 & 0.2481\\
\!\!\!{\footnotesize fc7+RBF-SVM}\!\!\! & 0.3666 & 0.2994 & 0.3351 & 0.2617\\
{\footnotesize fc7+NearestNeighbor} & 0.4195 & 0.3337 & 0.4159 & 0.3319\\
CNN-PDD & 0.2333 & 0.2205 & 0.3401 & 0.2188\\
\!\!\!{\footnotesize VGG-PDD-finetune}\!\!\! & 0.3269 & 0.2251 & 0.3198 & 0.2090\\
Fast-RCNN & 0.4131 & 0.3227 & 0.2245 & 0.2810\\
Ours & {\bf 0.1115} & {\bf 0.1388} & {\bf 0.0758} & {\bf 0.1368}\\
\hline
\end{tabular}}
\end{table}

\begin{table}[t]\caption{Part center prediction accuracy on the PASCAL VOC Part Dataset}
\label{tab:VOC}
\centering
\resizebox{1.0\linewidth}{!}{\begin{tabular}{c|p{1.8cm}|cccccc|c}
\hline
\multirow{5}{0.2cm}{\rotatebox{90}{Head}} & & bird & cat & cow & dog & horse & sheep & {\bf Avg.}\\
& SS-DPM &
9.0 & 39.4 & 39.0 & 49.4 & 38.0 & 36.8 & 35.3\\
& PL-DPM &
18.6 & 28.0 & 33.5 & 32.5 & 20.8 & 0 & 22.2\\
& PL-Graph &
11.9 & 53.6 & 34.6 & 56.4 & 20.1 & 0 & 29.4\\
& Fast-RCNN &
21.2 & 36.4 & 40.1 & 34.3 & 24.0 & 40.5 & 32.7\\
& Ours &
{\bf 64.5} & {\bf 85.0} & {\bf 65.4} & {\bf 81.8} & {\bf 77.1} & {\bf 64.1} & {\bf 73.0}\\
\hline
\multirow{5}{0.2cm}{\rotatebox{90}{Torso}} & & bird & cat & cow & dog & horse & sheep & {\bf Avg.}\\
& SS-DPM &
46.0 & 49.6 & 71.0 & 56.6 & 27.9 & 82.9 & 55.7\\
& PL-DPM &
68.0 & 38.0 & 78.2 & 20.2 & 75.5 & {\bf 92.8} & 62.1\\
& PL-Graph &
61.2 & 52.9 & 82.4 & 58.6 & 64.1 & 92.4 & 68.6\\
& Fast-RCNN &
52.2 & 70.3 & 86.5 & 43.8 & 57.6 & 91.3 & 66.9\\
& Ours &
{\bf 89.2} & {\bf 79.2} & {\bf 89.6} & {\bf 83.2} & {\bf 93.1} & 91.6 & {\bf 87.7}\\
\hline
\multicolumn{9}{c}{3-shot learning}\\
\hline
\multirow{5}{0.2cm}{\rotatebox{90}{Head}} & & bird & cat & cow & dog & horse & sheep & {\bf Avg.}\\
& SS-DPM &
13.5 & 47.8 & 30.8 & 51.8 & 38.0 & 30.2 & 35.3\\
& PL-DPM &
23.3 & 54.5 & 31.3 & 53.7 & 16.5 & 28.1 & 34.6\\
& PL-Graph &
21.0 & 34.2 & 25.3 & 55.6 & 25.8 & 22.3 & 30.7\\
& Fast-RCNN &
44.0 & 61.3 & 53.3 & 60.8 & 48.0 & 59.5 & 54.5\\
& Ours &
{\bf 68.7} & {\bf 85.5} & {\bf 67.0} & {\bf 78.6} & {\bf 79.9} & {\bf 71.1} & {\bf 75.1}\\
\hline
\multirow{5}{0.2cm}{\rotatebox{90}{Torso}} & & bird & cat & cow & dog & horse & sheep & {\bf Avg.}\\
& SS-DPM &
54.0 & 51.5 & 66.8 & 33.7 & 50.7 & 88.6 & 57.5\\
& PL-DPM &
68.7 & 65.0 & 77.2 & 29.8 & 77.2 & 89.0 & 67.8\\
& PL-Graph &
26.6 & 65.1 & 85.0 & 65.1 & 79.3 & 89.0 & 68.4\\
& Fast-RCNN &
70.5 & 76.3 & {\bf 90.7} & 68.9 & 70.7 & 93.5 & 78.4\\
& Ours &
{\bf 87.3} & {\bf 84.3} & 89.1 & {\bf 84.6} & {\bf 91.4} & {\bf 94.3} & {\bf 88.5}\\
\hline
\multicolumn{9}{c}{12-shot learning}
\end{tabular}}
\end{table}

\begin{figure}[t]
\centering
\includegraphics[width=0.9\linewidth]{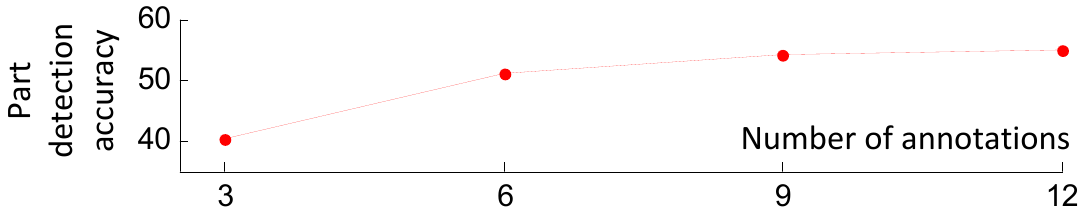}
\caption{Performance with different numbers of annotations. We annotate 1--4 parts for each of the 3 part templates.}
\label{fig:incremental}
\end{figure}

\subsection{Evaluation metric}

As mentioned in \cite{SemanticPart}, a fair evaluation of part localization requires to remove the factors of object detection. Therefore, we used object bounding boxes to crop objects from the original images as the testing samples. Note that detection-based baselines (\emph{e.g.} fast-RCNN, PL-Graph) may produce several bounding boxes for the part. Just as in \cite{SemanticPart,ObjectDiscoveryCNN_1}, we took the most confident bounding box per image as the localization result. Given localization results of a part in a certain category, we used three evaluation metrics. 1) \textit{Part detection}: a true part detection was identified based on the widely used ``IOU $\geq0.5$'' criterion~\cite{FastRCNN}; the part detection rate of this category was computed. 2) \textit{Center prediction}: as in \cite{ObjectDiscoveryCNN_1}, if the predicted part center was localized inside the true part bounding box, we considered it a correct center prediction; otherwise not. The average center prediction rate was computed among all objects in the category for evaluation. 3) The \textit{normalized distance} in \cite{CNNSemanticPart} is a standard metric to evaluate localization accuracy on the CUB200-2011 dataset. Because object parts may not have clear boundaries (\emph{e.g.} the forehead of the bird), \textit{center prediction} and \textit{normalized distance} are more often used for evaluation of part localization.

\begin{table*}[t]
\caption{Part center prediction accuracy of 3-shot learning on the ILSVRC 2013 DET Animal-Part dataset.}
\label{tab:LOC}
\centering
\resizebox{1.0\linewidth}{!}{\begin{tabular}{c|p{5.6cm}|cccccccccccccccc}
\toprule
\multirow{12}{0.2cm}{\rotatebox{90}{Head}} \!\!\!\!&\!\!\! \!\!\!\!&\!\!\! bird \!\!\!\!&\!\!\! frog \!\!\!\!&\!\!\! turt. \!\!\!\!&\!\!\! liza. \!\!\!\!&\!\!\! koala \!\!\!\!&\!\!\! lobs. \!\!\!\!&\!\!\! dog \!\!\!\!&\!\!\! fox \!\!\!\!&\!\!\! cat \!\!\!\!&\!\!\! lion \!\!\!\!&\!\!\! tiger \!\!\!\!&\!\!\! bear \!\!\!\!&\!\!\! rabb. \!\!\!\!&\!\!\! hams. \!\!\!\!&\!\!\! squi. \!\!\!\!&\!\!\! horse\\
\!\!\!\!&\!\!\! SS-DPM~\cite{SSDPM}
\!\!\!\!&\!\!\!24.4
\!\!\!\!&\!\!\!31.8
\!\!\!\!&\!\!\!24.7
\!\!\!\!&\!\!\!20.7
\!\!\!\!&\!\!\!54.4
\!\!\!\!&\!\!\!25.5
\!\!\!\!&\!\!\!51.2
\!\!\!\!&\!\!\!32.7
\!\!\!\!&\!\!\!43.0
\!\!\!\!&\!\!\!51.0
\!\!\!\!&\!\!\!76.5
\!\!\!\!&\!\!\!46.0
\!\!\!\!&\!\!\!38.8
\!\!\!\!&\!\!\!76.3
\!\!\!\!&\!\!\!27.4
\!\!\!\!&\!\!\!34.7
\\
\!\!\!\!&\!\!\! PL-DPM~\cite{PLDPM}
\!\!\!\!&\!\!\!26.4
\!\!\!\!&\!\!\!19.4
\!\!\!\!&\!\!\!13.6
\!\!\!\!&\!\!\!14.8
\!\!\!\!&\!\!\!46.0
\!\!\!\!&\!\!\!25.5
\!\!\!\!&\!\!\!54.0
\!\!\!\!&\!\!\!0
\!\!\!\!&\!\!\!63.2
\!\!\!\!&\!\!\!52.6
\!\!\!\!&\!\!\!75.5
\!\!\!\!&\!\!\!29.0
\!\!\!\!&\!\!\!56.8
\!\!\!\!&\!\!\!80.5
\!\!\!\!&\!\!\!20.9
\!\!\!\!&\!\!\!35.3
\\
\!\!\!\!&\!\!\! PL-Graph~\cite{SemanticPart}
\!\!\!\!&\!\!\!25.4
\!\!\!\!&\!\!\!43.5
\!\!\!\!&\!\!\!18.8
\!\!\!\!&\!\!\!16.3
\!\!\!\!&\!\!\!56.7
\!\!\!\!&\!\!\!27.0
\!\!\!\!&\!\!\!56.9
\!\!\!\!&\!\!\!37.4
\!\!\!\!&\!\!\!52.3
\!\!\!\!&\!\!\!67.0
\!\!\!\!&\!\!\!77.0
\!\!\!\!&\!\!\!48.0
\!\!\!\!&\!\!\!63.9
\!\!\!\!&\!\!\!84.7
\!\!\!\!&\!\!\!45.6
\!\!\!\!&\!\!\!39.5
\\
\!\!\!\!&\!\!\! Fast-RCNN~\cite{FastRCNN}
\!\!\!\!&\!\!\!24.4
\!\!\!\!&\!\!\!11.2
\!\!\!\!&\!\!\!24.7
\!\!\!\!&\!\!\!5.9
\!\!\!\!&\!\!\!79.5
\!\!\!\!&\!\!\!9.2
\!\!\!\!&\!\!\!21.3
\!\!\!\!&\!\!\!62.1
\!\!\!\!&\!\!\!63.7
\!\!\!\!&\!\!\!75.3
\!\!\!\!&\!\!\!69.9
\!\!\!\!&\!\!\!49.5
\!\!\!\!&\!\!\!60.1
\!\!\!\!&\!\!\!71.6
\!\!\!\!&\!\!\!57.2
\!\!\!\!&\!\!\!16.3
\\
\!\!\!\!&\!\!\! Ours
\!\!\!\!&\!\!\!{\bf 88.6}
\!\!\!\!&\!\!\!{\bf 84.7}
\!\!\!\!&\!\!\!{\bf 57.8}
\!\!\!\!&\!\!\!{\bf 84.4}
\!\!\!\!&\!\!\!{\bf 95.8}
\!\!\!\!&\!\!\!{\bf 41.1}
\!\!\!\!&\!\!\!{\bf 90.5}
\!\!\!\!&\!\!\!{\bf 94.3}
\!\!\!\!&\!\!\!{\bf 81.3}
\!\!\!\!&\!\!\!{\bf 87.6}
\!\!\!\!&\!\!\!{\bf 92.3}
\!\!\!\!&\!\!\!{\bf 84.0}
\!\!\!\!&\!\!\!{\bf 96.2}
\!\!\!\!&\!\!\!{\bf 96.7}
\!\!\!\!&\!\!\!{\bf 90.2}
\!\!\!\!&\!\!\!{\bf 74.2}
\\
\!\!\!\!&\!\!\! \!\!\!\!&\!\!\! zebra \!\!\!\!&\!\!\! swine \!\!\!\!&\!\!\! hippo \!\!\!\!&\!\!\! catt. \!\!\!\!&\!\!\! sheep \!\!\!\!&\!\!\! ante. \!\!\!\!&\!\!\! camel \!\!\!\!&\!\!\! otter \!\!\!\!&\!\!\! arma. \!\!\!\!&\!\!\! monk. \!\!\!\!&\!\!\! elep. \!\!\!\!&\!\!\! red pa. \!\!\!\!&\!\!\! gia.pa. \!\!\!\!&\!\!\! gold. \!\!\!\!&\!\!\! \!\!\!\!&\!\!\! \textcolor{blue}{\bf\normalsize Avg.}\\
\!\!\!\!&\!\!\! SS-DPM~\cite{SSDPM}
\!\!\!\!&\!\!\!42.9
\!\!\!\!&\!\!\!56.7
\!\!\!\!&\!\!\!64.4
\!\!\!\!&\!\!\!50.8
\!\!\!\!&\!\!\!52.1
\!\!\!\!&\!\!\!48.8
\!\!\!\!&\!\!\!35.3
\!\!\!\!&\!\!\!36.9
\!\!\!\!&\!\!\!28.3
\!\!\!\!&\!\!\!57.0
\!\!\!\!&\!\!\!73.0
\!\!\!\!&\!\!\!50.9
\!\!\!\!&\!\!\!77.7
\!\!\!\!&\!\!\!29.0
\!\!\!\!&\!\!\!
\!\!\!\!&\!\!\!\textcolor{blue}{45.4}
\\
\!\!\!\!&\!\!\! PL-DPM~\cite{PLDPM}
\!\!\!\!&\!\!\!30.5
\!\!\!\!&\!\!\!48.2
\!\!\!\!&\!\!\!53.2
\!\!\!\!&\!\!\!47.0
\!\!\!\!&\!\!\!7.4
\!\!\!\!&\!\!\!17.1
\!\!\!\!&\!\!\!39.5
\!\!\!\!&\!\!\!32.1
\!\!\!\!&\!\!\!21.2
\!\!\!\!&\!\!\!59.7
\!\!\!\!&\!\!\!61.2
\!\!\!\!&\!\!\!49.6
\!\!\!\!&\!\!\!1.8
\!\!\!\!&\!\!\!18.0
\!\!\!\!&\!\!\!
\!\!\!\!&\!\!\!\textcolor{blue}{36.7}
\\
\!\!\!\!&\!\!\! PL-Graph~\cite{SemanticPart}
\!\!\!\!&\!\!\!36.2
\!\!\!\!&\!\!\!43.3
\!\!\!\!&\!\!\!42.6
\!\!\!\!&\!\!\!38.9
\!\!\!\!&\!\!\!47.2
\!\!\!\!&\!\!\!46.5
\!\!\!\!&\!\!\!38.6
\!\!\!\!&\!\!\!48.7
\!\!\!\!&\!\!\!26.3
\!\!\!\!&\!\!\!59.7
\!\!\!\!&\!\!\!59.9
\!\!\!\!&\!\!\!42.5
\!\!\!\!&\!\!\!84.1
\!\!\!\!&\!\!\!17.0
\!\!\!\!&\!\!\!
\!\!\!\!&\!\!\!\textcolor{blue}{46.4}
\\
\!\!\!\!&\!\!\! Fast-RCNN~\cite{FastRCNN}
\!\!\!\!&\!\!\!25.4
\!\!\!\!&\!\!\!53.7
\!\!\!\!&\!\!\!58.5
\!\!\!\!&\!\!\!56.8
\!\!\!\!&\!\!\!59.5
\!\!\!\!&\!\!\!64.5
\!\!\!\!&\!\!\!41.4
\!\!\!\!&\!\!\!43.9
\!\!\!\!&\!\!\!31.8
\!\!\!\!&\!\!\!52.2
\!\!\!\!&\!\!\!77.6
\!\!\!\!&\!\!\!56.6
\!\!\!\!&\!\!\!65.0
\!\!\!\!&\!\!\!18.5
\!\!\!\!&\!\!\!
\!\!\!\!&\!\!\!\textcolor{blue}{46.9}
\\
\!\!\!\!&\!\!\! Ours
\!\!\!\!&\!\!\!{\bf 80.8}
\!\!\!\!&\!\!\!{\bf 73.2}
\!\!\!\!&\!\!\!{\bf 79.3}
\!\!\!\!&\!\!\!{\bf 80.0}
\!\!\!\!&\!\!\!{\bf 88.3}
\!\!\!\!&\!\!\!{\bf 93.1}
\!\!\!\!&\!\!\!{\bf 92.6}
\!\!\!\!&\!\!\!{\bf 88.8}
\!\!\!\!&\!\!\!{\bf 42.9}
\!\!\!\!&\!\!\!{\bf 94.1}
\!\!\!\!&\!\!\!{\bf 85.5}
\!\!\!\!&\!\!\!{\bf 80.7}
\!\!\!\!&\!\!\!{\bf 93.6}
\!\!\!\!&\!\!\!{\bf 83.0}
\!\!\!\!&\!\!\!
\!\!\!\!&\!\!\!\textcolor{blue}{{\bf 83.2}}
\\
\bottomrule
\multirow{12}{0.2cm}{\rotatebox{90}{Front legs}} \!\!\!\!&\!\!\! \!\!\!\!&\!\!\! bird \!\!\!\!&\!\!\! frog \!\!\!\!&\!\!\! turt. \!\!\!\!&\!\!\! liza. \!\!\!\!&\!\!\! koala \!\!\!\!&\!\!\! lobs. \!\!\!\!&\!\!\! dog \!\!\!\!&\!\!\! fox \!\!\!\!&\!\!\! cat \!\!\!\!&\!\!\! lion \!\!\!\!&\!\!\! tiger \!\!\!\!&\!\!\! bear \!\!\!\!&\!\!\! rabb. \!\!\!\!&\!\!\! hams. \!\!\!\!&\!\!\! squi. \!\!\!\!&\!\!\! horse\\
\!\!\!\!&\!\!\! SS-DPM~\cite{SSDPM}
\!\!\!\!&\!\!\!15.4
\!\!\!\!&\!\!\!41.8
\!\!\!\!&\!\!\!51.9
\!\!\!\!&\!\!\!33.0
\!\!\!\!&\!\!\!51.1
\!\!\!\!&\!\!\!63.1
\!\!\!\!&\!\!\!31.2
\!\!\!\!&\!\!\!35.7
\!\!\!\!&\!\!\!28.5
\!\!\!\!&\!\!\!38.6
\!\!\!\!&\!\!\!51.8
\!\!\!\!&\!\!\!52.2
\!\!\!\!&\!\!\!25.3
\!\!\!\!&\!\!\!43.8
\!\!\!\!&\!\!\!27.1
\!\!\!\!&\!\!\!33.8
\\
\!\!\!\!&\!\!\! PL-DPM~\cite{PLDPM}
\!\!\!\!&\!\!\!10.5
\!\!\!\!&\!\!\!31.6
\!\!\!\!&\!\!\!46.8
\!\!\!\!&\!\!\!8.3
\!\!\!\!&\!\!\!47.8
\!\!\!\!&\!\!\!64.1
\!\!\!\!&\!\!\!24.6
\!\!\!\!&\!\!\!14.7
\!\!\!\!&\!\!\!13.1
\!\!\!\!&\!\!\!34.6
\!\!\!\!&\!\!\!18.4
\!\!\!\!&\!\!\!37.7
\!\!\!\!&\!\!\!2.7
\!\!\!\!&\!\!\!25.8
\!\!\!\!&\!\!\!18.8
\!\!\!\!&\!\!\!33.8
\\
\!\!\!\!&\!\!\! PL-Graph~\cite{SemanticPart}
\!\!\!\!&\!\!\!0
\!\!\!\!&\!\!\!14.3
\!\!\!\!&\!\!\!54.4
\!\!\!\!&\!\!\!14.7
\!\!\!\!&\!\!\!38.0
\!\!\!\!&\!\!\!57.3
\!\!\!\!&\!\!\!23.2
\!\!\!\!&\!\!\!21.0
\!\!\!\!&\!\!\!23.8
\!\!\!\!&\!\!\!27.6
\!\!\!\!&\!\!\!32.6
\!\!\!\!&\!\!\!32.6
\!\!\!\!&\!\!\!17.3
\!\!\!\!&\!\!\!35.2
\!\!\!\!&\!\!\!27.1
\!\!\!\!&\!\!\!27.1
\\
\!\!\!\!&\!\!\! Fast-RCNN~\cite{FastRCNN}
\!\!\!\!&\!\!\!30.2
\!\!\!\!&\!\!\!29.6
\!\!\!\!&\!\!\!43.0
\!\!\!\!&\!\!\!7.3
\!\!\!\!&\!\!\!62.0
\!\!\!\!&\!\!\!60.2
\!\!\!\!&\!\!\!21.7
\!\!\!\!&\!\!\!23.8
\!\!\!\!&\!\!\!24.6
\!\!\!\!&\!\!\!28.3
\!\!\!\!&\!\!\!46.8
\!\!\!\!&\!\!\!57.2
\!\!\!\!&\!\!\!16.0
\!\!\!\!&\!\!\!24.2
\!\!\!\!&\!\!\!8.2
\!\!\!\!&\!\!\!34.6
\\
\!\!\!\!&\!\!\! Ours
\!\!\!\!&\!\!\!{\bf 58.6}
\!\!\!\!&\!\!\!{\bf 77.6}
\!\!\!\!&\!\!\!{\bf 67.1}
\!\!\!\!&\!\!\!{\bf 37.6}
\!\!\!\!&\!\!\!{\bf 82.6}
\!\!\!\!&\!\!\!{\bf 77.7}
\!\!\!\!&\!\!\!{\bf 85.5}
\!\!\!\!&\!\!\!{\bf 73.4}
\!\!\!\!&\!\!\!{\bf 62.3}
\!\!\!\!&\!\!\!{\bf 80.3}
\!\!\!\!&\!\!\!{\bf 77.3}
\!\!\!\!&\!\!\!{\bf 78.3}
\!\!\!\!&\!\!\!{\bf 66.7}
\!\!\!\!&\!\!\!{\bf 60.9}
\!\!\!\!&\!\!\!{\bf 63.8}
\!\!\!\!&\!\!\!{\bf 70.7}
\\
\!\!\!\!&\!\!\! \!\!\!\!&\!\!\! zebra \!\!\!\!&\!\!\! swine \!\!\!\!&\!\!\! hippo \!\!\!\!&\!\!\! catt. \!\!\!\!&\!\!\! sheep \!\!\!\!&\!\!\! ante. \!\!\!\!&\!\!\! camel \!\!\!\!&\!\!\! otter \!\!\!\!&\!\!\! arma. \!\!\!\!&\!\!\! monk. \!\!\!\!&\!\!\! elep. \!\!\!\!&\!\!\! red pa. \!\!\!\!&\!\!\! gia.pa. \!\!\!\!&\!\!\! gold. \!\!\!\!&\!\!\! \!\!\!\!&\!\!\! \textcolor{blue}{\bf\normalsize Avg.}\\
\!\!\!\!&\!\!\! SS-DPM~\cite{SSDPM}
\!\!\!\!&\!\!\!39.1
\!\!\!\!&\!\!\!28.8
\!\!\!\!&\!\!\!42.8
\!\!\!\!&\!\!\!42.6
\!\!\!\!&\!\!\!47.9
\!\!\!\!&\!\!\!59.6
\!\!\!\!&\!\!\!42.2
\!\!\!\!&\!\!\!38.9
\!\!\!\!&\!\!\!6.3
\!\!\!\!&\!\!\!54.8
\!\!\!\!&\!\!\!58.1
\!\!\!\!&\!\!\!40.3
\!\!\!\!&\!\!\!50.4
\!\!\!\!&\!\!\!-
\!\!\!\!&\!\!\!
\!\!\!\!&\!\!\!\textcolor{blue}{40.6}
\\
\!\!\!\!&\!\!\! PL-DPM~\cite{PLDPM}
\!\!\!\!&\!\!\!18.5
\!\!\!\!&\!\!\!16.1
\!\!\!\!&\!\!\!21.0
\!\!\!\!&\!\!\!0
\!\!\!\!&\!\!\!0
\!\!\!\!&\!\!\!33.1
\!\!\!\!&\!\!\!33.6
\!\!\!\!&\!\!\!10.5
\!\!\!\!&\!\!\!1.0
\!\!\!\!&\!\!\!48.7
\!\!\!\!&\!\!\!31.5
\!\!\!\!&\!\!\!14.5
\!\!\!\!&\!\!\!26.0
\!\!\!\!&\!\!\!-
\!\!\!\!&\!\!\!
\!\!\!\!&\!\!\!\textcolor{blue}{23.7}
\\
\!\!\!\!&\!\!\! PL-Graph~\cite{SemanticPart}
\!\!\!\!&\!\!\!17.9
\!\!\!\!&\!\!\!27.1
\!\!\!\!&\!\!\!23.2
\!\!\!\!&\!\!\!16.5
\!\!\!\!&\!\!\!54.2
\!\!\!\!&\!\!\!34.9
\!\!\!\!&\!\!\!33.6
\!\!\!\!&\!\!\!27.4
\!\!\!\!&\!\!\!12.5
\!\!\!\!&\!\!\!54.8
\!\!\!\!&\!\!\!54.8
\!\!\!\!&\!\!\!14.5
\!\!\!\!&\!\!\!38.2
\!\!\!\!&\!\!\!-
\!\!\!\!&\!\!\!
\!\!\!\!&\!\!\!\textcolor{blue}{29.5}
\\
\!\!\!\!&\!\!\! Fast-RCNN~\cite{FastRCNN}
\!\!\!\!&\!\!\!37.1
\!\!\!\!&\!\!\!40.7
\!\!\!\!&\!\!\!18.1
\!\!\!\!&\!\!\!39.1
\!\!\!\!&\!\!\!17.7
\!\!\!\!&\!\!\!29.5
\!\!\!\!&\!\!\!45.7
\!\!\!\!&\!\!\!17.9
\!\!\!\!&\!\!\!12.5
\!\!\!\!&\!\!\!29.6
\!\!\!\!&\!\!\!71.0
\!\!\!\!&\!\!\!25.8
\!\!\!\!&\!\!\!67.2
\!\!\!\!&\!\!\!-
\!\!\!\!&\!\!\!
\!\!\!\!&\!\!\!\textcolor{blue}{32.3}
\\
\!\!\!\!&\!\!\! Ours
\!\!\!\!&\!\!\!{\bf 83.4}
\!\!\!\!&\!\!\!{\bf 69.5}
\!\!\!\!&\!\!\!{\bf 59.4}
\!\!\!\!&\!\!\!{\bf 76.5}
\!\!\!\!&\!\!\!{\bf 77.1}
\!\!\!\!&\!\!\!{\bf 89.8}
\!\!\!\!&\!\!\!{\bf 87.9}
\!\!\!\!&\!\!\!{\bf 43.2}
\!\!\!\!&\!\!\!{\bf 61.5}
\!\!\!\!&\!\!\!{\bf 75.7}
\!\!\!\!&\!\!\!{\bf 84.7}
\!\!\!\!&\!\!\!{\bf 61.3}
\!\!\!\!&\!\!\!{\bf 74.8}
\!\!\!\!&\!\!\!-
\!\!\!\!&\!\!\!
\!\!\!\!&\!\!\!\textcolor{blue}{{\bf 71.2}}
\\
\bottomrule
\end{tabular}}
\end{table*}

\begin{table*}[t]
\caption{Part center prediction accuracy of 12-shot learning on the ILSVRC 2013 DET Animal-Part dataset.}
\label{tab:LOC12}
\centering
\resizebox{1.0\linewidth}{!}{\begin{tabular}{c|p{5.6cm}|cccccccccccccccc}
\toprule
\multirow{12}{0.2cm}{\rotatebox{90}{Head}} \!\!\!\!&\!\!\! \!\!\!\!&\!\!\! bird \!\!\!\!&\!\!\! frog \!\!\!\!&\!\!\! turt. \!\!\!\!&\!\!\! liza. \!\!\!\!&\!\!\! koala \!\!\!\!&\!\!\! lobs. \!\!\!\!&\!\!\! dog \!\!\!\!&\!\!\! fox \!\!\!\!&\!\!\! cat \!\!\!\!&\!\!\! lion \!\!\!\!&\!\!\! tiger \!\!\!\!&\!\!\! bear \!\!\!\!&\!\!\! rabb. \!\!\!\!&\!\!\! hams. \!\!\!\!&\!\!\! squi. \!\!\!\!&\!\!\! horse\\
\!\!\!\!&\!\!\! SS-DPM~\cite{SSDPM}
\!\!\!\!&\!\!\!29.9
\!\!\!\!&\!\!\!51.8
\!\!\!\!&\!\!\!26.0
\!\!\!\!&\!\!\!28.9
\!\!\!\!&\!\!\!72.6
\!\!\!\!&\!\!\!27.7
\!\!\!\!&\!\!\!53.1
\!\!\!\!&\!\!\!66.4
\!\!\!\!&\!\!\!64.2
\!\!\!\!&\!\!\!76.3
\!\!\!\!&\!\!\!82.1
\!\!\!\!&\!\!\!68.0
\!\!\!\!&\!\!\!66.1
\!\!\!\!&\!\!\!81.9
\!\!\!\!&\!\!\!37.7
\!\!\!\!&\!\!\!27.9
\\
\!\!\!\!&\!\!\! PL-DPM~\cite{PLDPM}
\!\!\!\!&\!\!\!11.9
\!\!\!\!&\!\!\!32.4
\!\!\!\!&\!\!\!24.0
\!\!\!\!&\!\!\!17.8
\!\!\!\!&\!\!\!55.3
\!\!\!\!&\!\!\!22.0
\!\!\!\!&\!\!\!37.9
\!\!\!\!&\!\!\!46.9
\!\!\!\!&\!\!\!53.4
\!\!\!\!&\!\!\!62.9
\!\!\!\!&\!\!\!67.9
\!\!\!\!&\!\!\!53.0
\!\!\!\!&\!\!\!67.2
\!\!\!\!&\!\!\!76.3
\!\!\!\!&\!\!\!22.3
\!\!\!\!&\!\!\!35.3
\\
\!\!\!\!&\!\!\! PL-Graph~\cite{SemanticPart}
\!\!\!\!&\!\!\!37.3
\!\!\!\!&\!\!\!27.1
\!\!\!\!&\!\!\!10.4
\!\!\!\!&\!\!\!21.5
\!\!\!\!&\!\!\!55.8
\!\!\!\!&\!\!\!5.0
\!\!\!\!&\!\!\!68.7
\!\!\!\!&\!\!\!38.9
\!\!\!\!&\!\!\!75.6
\!\!\!\!&\!\!\!77.3
\!\!\!\!&\!\!\!63.3
\!\!\!\!&\!\!\!55.5
\!\!\!\!&\!\!\!67.2
\!\!\!\!&\!\!\!69.8
\!\!\!\!&\!\!\!36.7
\!\!\!\!&\!\!\!28.4
\\
\!\!\!\!&\!\!\! Fast-RCNN~\cite{FastRCNN}
\!\!\!\!&\!\!\!55.7
\!\!\!\!&\!\!\!37.6
\!\!\!\!&\!\!\!58.4
\!\!\!\!&\!\!\!63.7
\!\!\!\!&\!\!\!85.1
\!\!\!\!&\!\!\!9.2
\!\!\!\!&\!\!\!54.5
\!\!\!\!&\!\!\!73.0
\!\!\!\!&\!\!\!64.2
\!\!\!\!&\!\!\!90.2
\!\!\!\!&\!\!\!84.7
\!\!\!\!&\!\!\!58.0
\!\!\!\!&\!\!\!85.2
\!\!\!\!&\!\!\!82.8
\!\!\!\!&\!\!\!78.1
\!\!\!\!&\!\!\!44.7
\\
\!\!\!\!&\!\!\! Ours
\!\!\!\!&\!\!\!{\bf 87.1}
\!\!\!\!&\!\!\!{\bf 86.5}
\!\!\!\!&\!\!\!{\bf 59.1}
\!\!\!\!&\!\!\!{\bf 80.0}
\!\!\!\!&\!\!\!{\bf 94.9}
\!\!\!\!&\!\!\!{\bf 44.7}
\!\!\!\!&\!\!\!{\bf 88.2}
\!\!\!\!&\!\!\!{\bf 88.6}
\!\!\!\!&\!\!\!{\bf 83.9}
\!\!\!\!&\!\!\!{\bf 90.2}
\!\!\!\!&\!\!\!{\bf 89.8}
\!\!\!\!&\!\!\!{\bf 91.5}
\!\!\!\!&\!\!\!{\bf 90.7}
\!\!\!\!&\!\!\!{\bf 97.7}
\!\!\!\!&\!\!\!{\bf 94.4}
\!\!\!\!&\!\!\!{\bf 72.6}
\\
\!\!\!\!&\!\!\! \!\!\!\!&\!\!\! zebra \!\!\!\!&\!\!\! swine \!\!\!\!&\!\!\! hippo \!\!\!\!&\!\!\! catt. \!\!\!\!&\!\!\! sheep \!\!\!\!&\!\!\! ante. \!\!\!\!&\!\!\! camel \!\!\!\!&\!\!\! otter \!\!\!\!&\!\!\! arma. \!\!\!\!&\!\!\! monk. \!\!\!\!&\!\!\! elep. \!\!\!\!&\!\!\! red pa. \!\!\!\!&\!\!\! gia.pa. \!\!\!\!&\!\!\! gold. \!\!\!\!&\!\!\! \!\!\!\!&\!\!\! \textcolor{blue}{\bf\normalsize Avg.}\\
\!\!\!\!&\!\!\! SS-DPM~\cite{SSDPM}
\!\!\!\!&\!\!\!46.3
\!\!\!\!&\!\!\!67.1
\!\!\!\!&\!\!\!61.7
\!\!\!\!&\!\!\!52.4
\!\!\!\!&\!\!\!66.3
\!\!\!\!&\!\!\!60.4
\!\!\!\!&\!\!\!41.4
\!\!\!\!&\!\!\!49.2
\!\!\!\!&\!\!\!42.4
\!\!\!\!&\!\!\!62.9
\!\!\!\!&\!\!\!78.3
\!\!\!\!&\!\!\!71.1
\!\!\!\!&\!\!\!87.7
\!\!\!\!&\!\!\!31.5
\!\!\!\!&\!\!\!
\!\!\!\!&\!\!\!\textcolor{blue}{56.0}
\\
\!\!\!\!&\!\!\! PL-DPM~\cite{PLDPM}
\!\!\!\!&\!\!\!34.5
\!\!\!\!&\!\!\!51.8
\!\!\!\!&\!\!\!54.3
\!\!\!\!&\!\!\!42.7
\!\!\!\!&\!\!\!35.6
\!\!\!\!&\!\!\!30.9
\!\!\!\!&\!\!\!33.0
\!\!\!\!&\!\!\!46.5
\!\!\!\!&\!\!\!15.7
\!\!\!\!&\!\!\!48.9
\!\!\!\!&\!\!\!60.5
\!\!\!\!&\!\!\!45.2
\!\!\!\!&\!\!\!74.5
\!\!\!\!&\!\!\!19.0
\!\!\!\!&\!\!\!
\!\!\!\!&\!\!\!\textcolor{blue}{42.7}
\\
\!\!\!\!&\!\!\! PL-Graph~\cite{SemanticPart}
\!\!\!\!&\!\!\!48.0
\!\!\!\!&\!\!\!43.3
\!\!\!\!&\!\!\!72.3
\!\!\!\!&\!\!\!36.8
\!\!\!\!&\!\!\!50.3
\!\!\!\!&\!\!\!41.5
\!\!\!\!&\!\!\!54.4
\!\!\!\!&\!\!\!60.4
\!\!\!\!&\!\!\!34.3
\!\!\!\!&\!\!\!66.7
\!\!\!\!&\!\!\!57.2
\!\!\!\!&\!\!\!68.0
\!\!\!\!&\!\!\!86.4
\!\!\!\!&\!\!\!18.5
\!\!\!\!&\!\!\!
\!\!\!\!&\!\!\!\textcolor{blue}{49.2}
\\
\!\!\!\!&\!\!\! Fast-RCNN~\cite{FastRCNN}
\!\!\!\!&\!\!\!{\bf62.1}
\!\!\!\!&\!\!\!67.7
\!\!\!\!&\!\!\!{\bf88.8}
\!\!\!\!&\!\!\!76.2
\!\!\!\!&\!\!\!76.1
\!\!\!\!&\!\!\!83.9
\!\!\!\!&\!\!\!70.7
\!\!\!\!&\!\!\!66.3
\!\!\!\!&\!\!\!65.2
\!\!\!\!&\!\!\!73.1
\!\!\!\!&\!\!\!83.6
\!\!\!\!&\!\!\!83.3
\!\!\!\!&\!\!\!79.1
\!\!\!\!&\!\!\!86.5
\!\!\!\!&\!\!\!
\!\!\!\!&\!\!\!\textcolor{blue}{69.6}
\\
\!\!\!\!&\!\!\! Ours
\!\!\!\!&\!\!\!51.4
\!\!\!\!&\!\!\!81.7
\!\!\!\!&\!\!\!{\bf 82.4}
\!\!\!\!&\!\!\!{\bf 82.4}
\!\!\!\!&\!\!\!{\bf 87.1}
\!\!\!\!&\!\!\!{\bf 87.1}
\!\!\!\!&\!\!\!{\bf 91.2}
\!\!\!\!&\!\!\!{\bf 92.5}
\!\!\!\!&\!\!\!{\bf 74.7}
\!\!\!\!&\!\!\!{\bf 93.7}
\!\!\!\!&\!\!\!{\bf 90.1}
\!\!\!\!&\!\!\!{\bf 92.1}
\!\!\!\!&\!\!\!{\bf 93.2}
\!\!\!\!&\!\!\!{\bf 95.0}
\!\!\!\!&\!\!\!
\!\!\!\!&\!\!\!\textcolor{blue}{{\bf 84.5}}
\\
\hline
\end{tabular}}
\end{table*}

\begin{figure*}[t]
\centering
\includegraphics[width=0.99\linewidth]{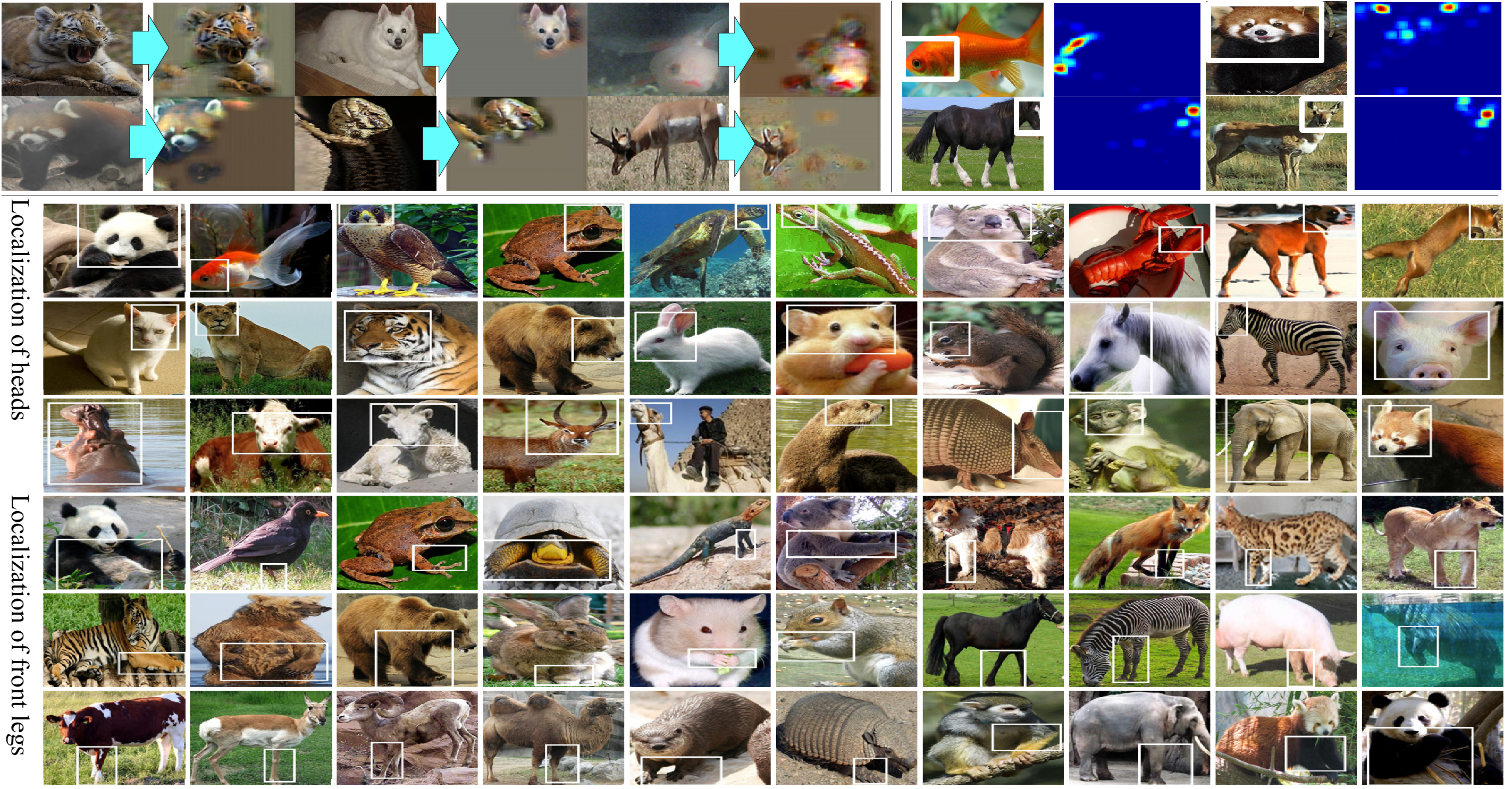}
\caption{Image reconstruction based on the AOG (top left), heat maps corresponding to latent patterns at the 5-th conv-layers in the VGG-16 network (top right), and part localization performance to demonstrate the AOG interpretability (bottom).}
\label{fig:perf}
\end{figure*}

\subsection{Results and quantitative analysis}

Table~\ref{tab:stat} lists the average children number of an AOG node at different layers. Fig.~\ref{fig:perf} shows the positions of the extracted latent pattern nodes, and part-localization results based on the AOGs. Given an image, we also used latent patterns in the AOG to reconstruct the corresponding semantic part based on the technique of \cite{FeaVisual}, in order to show the interpretability of the AOG.

In Tables~\ref{tab:IoU}, \ref{tab:dist}, \ref{tab:VOC}, \ref{tab:LOC}, and \ref{tab:LOC12}, we compared the performance of different baselines. Our method exhibited much better performance than other baselines that suffered from over-fitting problems. In Fig.~\ref{fig:incremental}, we showed the performance curve when we increased the annotation number from 3 to 12. Note that the 12-shot learning only improved about 0.9\%--2.9\% of center prediction over the 3-shot learning. This demonstrated that our method was efficient in mining CNN semantics, and the CNN units related to each part template had been roughly mined using just three annotations. In fact, we can further improve the performance by defining more part templates, rather than by annotating more part boxes for existing part templates.

\section{Conclusions and discussion}

In this paper, we have presented a method for incrementally growing new neural connections on a pre-trained CNN to encode new semantic parts in a four-layer AOG. Given an demand for modeling a semantic part on the fly, our method can be conducted with a small number of part annotations (even a single box annotation for each part template). In addition, our method semanticizes CNN units by associating them with certain semantic parts, and builds an AOG as a interpretable model to explain the semantic hierarchy of CNN units.

Because we reduce high-dimensional CNN activations to low-dimensional representation of parts/sub-parts, our method has high robustness and efficiency in multi-shot learning, and has exhibited superior performance to other baselines.

\section{Acknowledgement}

This study is supported by MURI project N00014-16-1-2007 and DARPA SIMPLEX project N66001-15-C-4035.

\section{Appendix}

\subsection{Parameters for latent patterns}

In general, we use the notation of ${\bf P}_{V}$ to denote the central position of an image region $\Lambda_{V}$ as follows.
\begin{center}
\begin{tabular}{lp{6.1cm}}
\hline
${\bf P}_{V^{\textrm{tmp}}}$ & center of ${\Lambda}_{V^{\textrm{tmp}}}$\\
$\hat{\bf P}_{V^{\textrm{lat}}}$ & center of $\hat{\Lambda}_{V^{\textrm{lat}}}$ obtained during part parsing\\
${\bf P}_{V^{\textrm{unt}}}$ & a constant center position of ${\Lambda}_{V^{\textrm{unt}}}$\\
$\overline{\bf P}_{V^{\textrm{lat}}}$ & an AOG parameter, the center of the square deformation range of $V^{\textrm{lat}}$, \emph{i.e.} $V^{\textrm{lat}}$'s ``ideal'' position without any deformation.\\
$\Delta{\bf P}_{V^{\textrm{lat}}}$ & an AOG parameter, average displacement from $V^{\textrm{lat}}$ to the parent $V^{\textrm{tmp}}$\\
\hline
\end{tabular}
\end{center}

Each latent pattern {\small$V^{\textrm{lat}}$} is defined by its location parameters {\small$\{L_{V^{\textrm{lat}}},D_{V^{\textrm{lat}}},\overline{\bf P}_{V^{\textrm{lat}}},\Delta{\bf P}_{V^{\textrm{lat}}}\}\subset{\boldsymbol\theta}$}, where {\small${\boldsymbol\theta}$} is the set of AOG parameters. It means that a latent pattern $V^{\textrm{lat}}$ uses a square\textcolor{red}{\footnotemark[5]} within the $D_{V^{\textrm{lat}}}$-th conv-slice/channel in the output of the $L_{V^{\textrm{lat}}}$-th CNN conv-layer as its deformation range. Each $V^{\textrm{lat}}$ in the $k$-th conv-layer has a fixed value of $L_{V^{\textrm{lat}}}=k$. $\Delta{\bf P}_{V^{\textrm{lat}}}$ is used to compute {\small$S^{\textrm{inf}}(\Lambda_{V^{\textrm{tmp}}}|\hat{\Lambda}_{V^{\textrm{lat}}})$}. Given parameter {\small$\overline{\bf P}_{V^{\textrm{lat}}}$}, the displacement {\small$\Delta{\bf P}_{V^{\textrm{lat}}}$} can be estimated as {\small$\Delta{\bf P}_{V^{\textrm{lat}}}\!=\!{\bf P}^{*}_{V^{\textrm{tmp}}}-\overline{\bf P}_{V^{\textrm{lat}}}$}, where {\small$\overline{\bf P}^{*}_{V^{\textrm{tmp}}}$} denotes the average position among all ground-truth parts that are annotated for {\small$V^{\textrm{tmp}}$}. As a result, for each latent pattern {\small$V^{\textrm{lat}}$}, we only need to learn its conv-slice {\small$D_{V^{\textrm{lat}}}\in{\boldsymbol\theta}$} and central position {\small$\overline{\bf P}_{V^{\textrm{lat}}}\in{\boldsymbol\theta}$}.

\subsection{Scores of terminal nodes}

The inference score for each terminal node $V^{\textrm{unt}}$ under a latent pattern $V^{\textrm{lat}}$ is formulated as
\begin{small}
\begin{eqnarray}
&S_{I}(V^{\textrm{unt}})=S_{I}^{\textrm{rsp}}(V^{\textrm{unt}})+S_{I}^{\textrm{loc}}(V^{\textrm{unt}})+S_{I}^{\textrm{pair}}(V^{\textrm{unt}})\nonumber\\
&S_{I}^{\textrm{rsp}}(V^{\textrm{unt}})=\left\{\begin{array}{ll}\lambda^{\textrm{rsp}}X(V^{\textrm{unt}}),& X(V^{\textrm{unt}})>0\\ \lambda^{\textrm{rsp}}S_{none},& X(V^{\textrm{unt}})\leq0\end{array}\right.\nonumber\\
&S_{I}^{\textrm{pair}}(V^{\textrm{unt}})=-\lambda^{\textrm{pair}}\!\!\!\!\!\!\!\!\!\!\!\!\underset{V^{\textrm{lat}}_{\textrm{upper}}\in\!\textrm{Neighbor}(V^{\textrm{lat}})}{\textrm{mean}}\!\!\!\!\!\!\!\!\!\!\!\Vert[{\bf P}_{V^{\textrm{unt}}}\!-\!\hat{\bf P}_{V^{\textrm{lat}}_{\textrm{upper}}}\!]\!-\![\overline{\bf P}_{V^{\textrm{lat}}_{\textrm{upper}}}\!\!-\!\overline{\bf P}_{V^{\textrm{lat}}}]\Vert\nonumber
\end{eqnarray}
\end{small}
The score of {\small$S_{I}(V^{\textrm{unt}})$} consists of the following three terms: 1) {\small$S_{I}^{\textrm{rsp}}(V^{\textrm{unt}})$} denotes the response value of the unit {\small$V^{\textrm{unt}}$}, when we input image $I$ into the CNN. $X(V^{\textrm{unt}})$ denotes the normalized response value of $V^{\textrm{unt}}$; $S_{none}=-3$ is set for non-activated units. 2) When the parent {\small$V^{\textrm{lat}}$} selects {\small$V^{\textrm{unt}}$} as its location inference (\emph{i.e.} {\small$\hat{\Lambda}_{V^{\textrm{lat}}}\leftarrow\Lambda_{V^{\textrm{unt}}}$}), {\small$S_{I}^{\textrm{loc}}(V^{\textrm{unt}})$} measures the deformation level between {\small$V^{\textrm{unt}}$}'s location {\small${\bf P}_{V^{\textrm{unt}}}$} and {\small$V^{\textrm{lat}}$}'s ideal location {\small$\overline{\bf P}_{V^{\textrm{lat}}}$}. 3) {\small$S_{I}^{\textrm{pair}}(V^{\textrm{unt}})$} indicates the spatial compatibility between neighboring latent patterns: we model the pairwise spatial relationship between latent patterns in the upper conv-layer and those in the current conv-layer. For each $V^{\textrm{unt}}$ (with its parent $V^{\textrm{lat}}$) in conv-layer $L_{V^{\textrm{lat}}}$, we select 15 nearest latent patterns in conv-layer $L_{V^{\textrm{lat}}}+1$, \emph{w.r.t.} {\small$\Vert\overline{\bf P}_{V^{\textrm{lat}}}-\overline{\bf P}_{V^{\textrm{lat}}_{\textrm{upper}}}\Vert$}, as the neighboring latent patterns. We set constant weights $\lambda^{\textrm{rsp}}=1.5,\lambda^{\textrm{loc}}=1/3,\lambda^{\textrm{pair}}=10.0$, $\lambda^{\textrm{unsup}}=5.0$, and $\lambda^{\textrm{close}}=0.4$ for all categories. Based on the above design, we first infer latent patterns corresponding to high conv-layers, and use the inference results to select units in low conv-layers.

\subsection{Scores of AND nodes}

\begin{small}
\begin{displaymath}
S^{\textrm{inf}}(\Lambda_{V^{\textrm{tmp}}}|\hat{\Lambda}_{V^{\textrm{lat}}})\!=\!-\lambda^{\textrm{inf}}\min\{\Vert\hat{\bf P}_{V^{\textrm{lat}}}+\Delta{\bf P}_{V^{\textrm{lat}}}-{\bf P}_{V^{\textrm{tmp}}}\Vert^2,d^2\}
\end{displaymath}
\end{small}
where we set $d\!=\!37$ pxls and $\lambda^{\textrm{inf}}=5.0$.

\bibliography{TheBib}
\bibliographystyle{aaai}

\end{document}